\documentclass[sigconf,screen,nonacm]{acmart}

\begin{document}

\title{Creating Visual Effects with Neural Radiance Fields}

\author{Cyrus Vachha}
\email{cvachha@berkeley.edu}

\affiliation{%
  \institution{University of California, Berkeley}
  \city{Berkeley}
  \state{California}
  \country{USA}
  \postcode{94704}
}

\ccsdesc[500]{Computing methodologies~Animation}
\ccsdesc[500]{Computing methodologies~Rendering}

\keywords{NeRFs, visual effects, animation, compositing, art}

\begin{teaserfigure}
  \includegraphics[width=\textwidth]{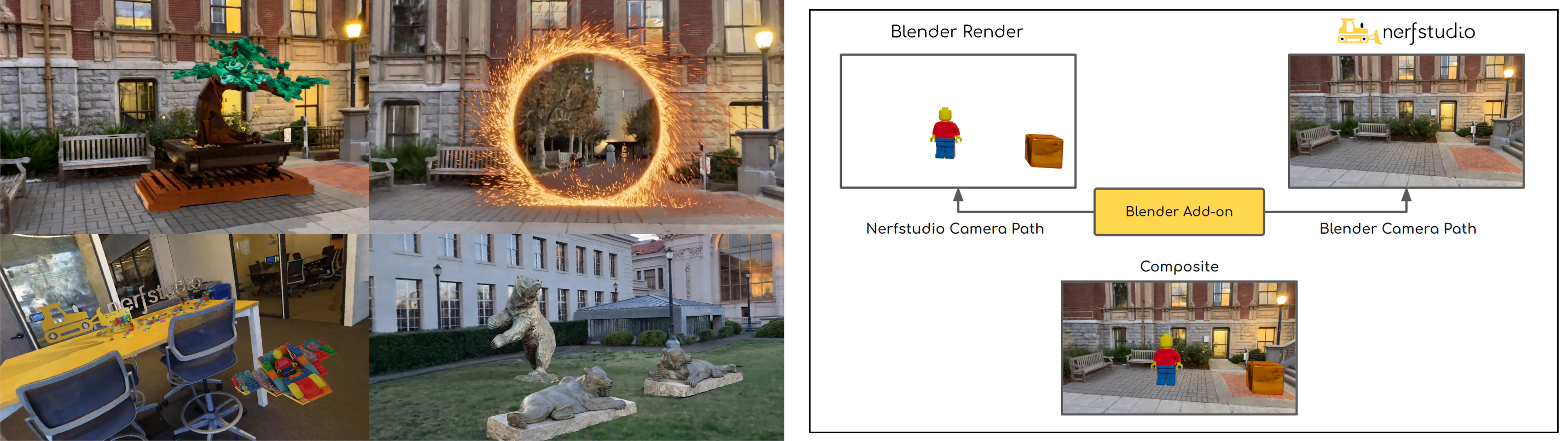}
  \caption{(Left) Renders created using the Nerfstudio Blender VFX add-on. (Right) Overview of the NeRF and Blender render compositing process}
  \Description{The renders show composites of NeRFs with meshes and with other NeRFs. The outline shows how the add-on helps render the NeRF and Blender camera to composite.}
  \label{fig:teaser}
\end{teaserfigure}

\maketitle

\section{Introduction}
Neural Radiance Fields (NeRFs) \cite{NeRF} have emerged as a popular research area in graphics for constructing 3D environments and objects. Capabilities of NeRFs, including 3D mesh exporting, photorealistic reconstruction, and rendering depth maps \cite{NeRF}, demonstrate how NeRFs can be incorporated into visual effects (VFX). Although current research on NeRFs is primarily focused on enhancing algorithms and improving quality, there has been less exploration of their potential in the field of VFX. Recent developments in toolkits for training and rendering NeRFs such as Nerfstudio \cite{tancik2023nerfstudio} and Instant NGP \cite{InstantNeRF}  have made NeRFs more accessible, leading to their increasing popularity and adoption by non-researchers in fields such as architecture, VFX, and virtual production. Startups such as Luma AI\footnote{https://lumalabs.ai/} and Volinga\footnote{https://volinga.ai/} offer Unreal Engine Integrations that allows NeRFs to be rendered in real-time within the editor. These integrations allow VFX artists to incorporate NeRFs environments and objects for virtual production, allowing for new forms of VFX and the creation of realistic and immersive environments. These solutions are designed for real-time rendered content within the Unreal Engine, but do not offer a pipeline into other commonly used visual effects programs and compositing pipelines.
We present a method for integrating novel 3D representations, such as NeRFs or 3D Gaussian Splatting \cite{kerbl3Dgaussians}, into traditional compositing VFX pipelines using Nerfstudio, an open-source framework for training and rendering NeRFs \cite{tancik2023nerfstudio}. Our approach involves using Blender\footnote{https://www.blender.org/}, a widely used 3D creation software, to align camera paths and composite NeRF renders with meshes and other NeRFs, allowing for seamless integration of NeRFs into traditional VFX pipelines. Blender is an industry used open-source 3D creation software that allows users to create, animate, and render 3D models and visual effects. Our Blender add-on script allows for more controlled camera trajectories of photorealistic scenes, compositing meshes and other environmental effects with NeRFs, and compositing multiple NeRFs in a single scene.
Documentation can be found here:
\url{https://docs.nerf.studio/extensions/blender_addon.html}.

\section{Implementation}
\begin{figure}[h]
  \centering
  \includegraphics[width=\linewidth]{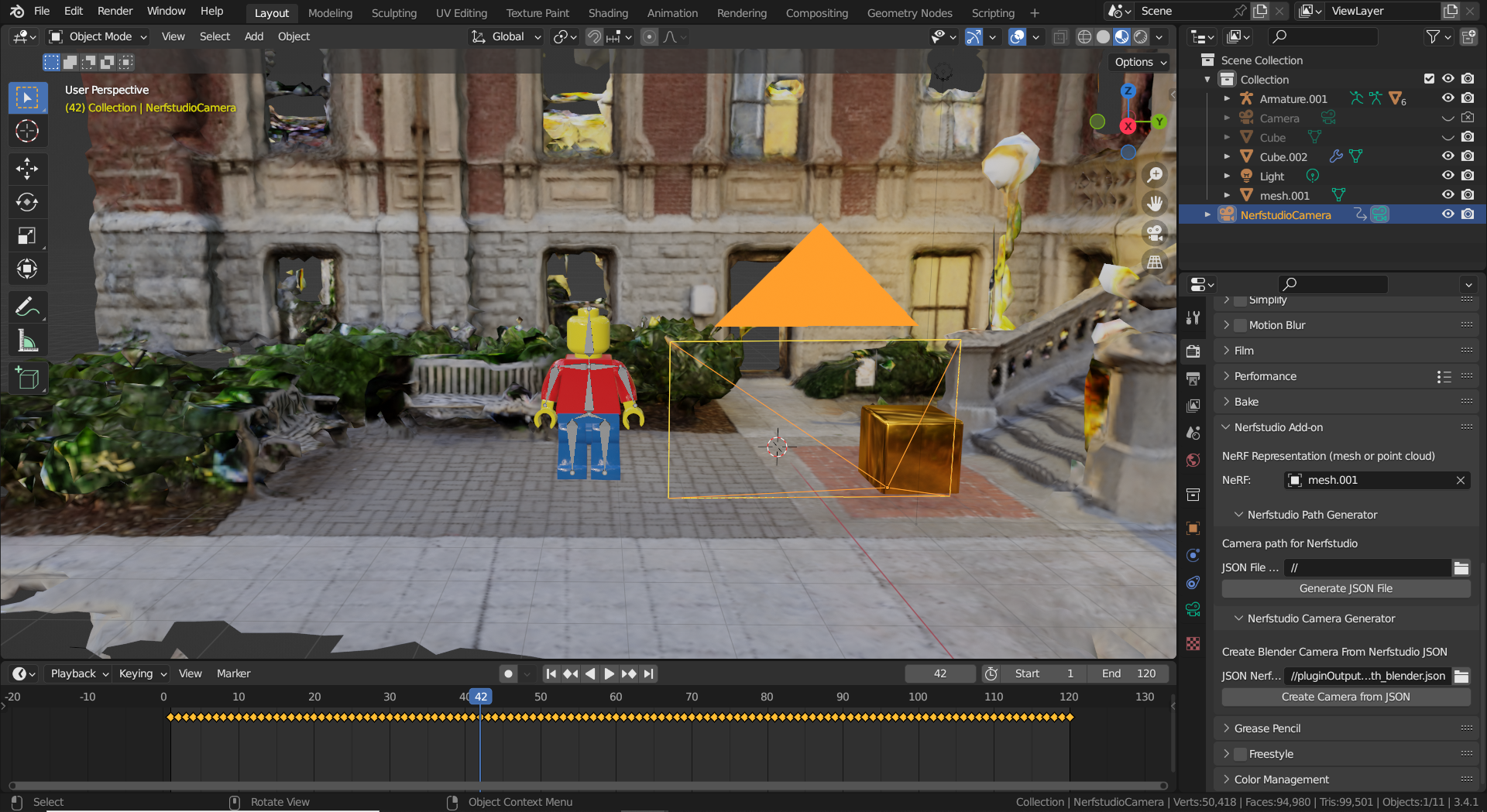}
  \caption{Blender editor with a NeRF representation and meshes using the add-on}
  \Description{Blender editor showing meshes in front of the NeRF mesh representation, with the add-on shown on the side panel.}
  \label{fig:blender-ui}
\end{figure}

We propose a method for integrating NeRFs into VFX by aligning NeRF renders with the virtual camera with meshes or other NeRF renders, and we have developed a Blender add-on to facilitate this process as shown in Figure~\ref{fig:teaser}. In our proposed method, NeRF representations can be imported into 3D creation tools, such as Blender, where the camera paths of the NeRF renders are generated to align with the virtual camera in Blender. By aligning the NeRF camera path with the virtual Blender camera, the NeRF renders can be integrated and composited with the Blender renders. A user simply selects the NeRF representation in the scene and our script will generate an aligned JSON camera path used to render the NeRF in Nerfstudio. A mesh or point cloud NeRF representation can be exported from the Nerfstudio editor and acts as a reference representation of the NeRF within Blender, which could contain additional NeRF representations or meshes as shown in Figure~\ref{fig:blender-ui}.

This is achieved by transforming the Blender camera path coordinate system to be relative to the origin of the NeRF representation in the Blender scene for each frame in the render. To align the position, rotation, and scale of the NeRF representation with its render, we iterate over the scene frame sequence and get the Blender camera’s 4x4 worldspace transformation matrix. We then obtain the worldspace matrix of the NeRF representation at each frame and transform the camera coordinates to the coordinate system of the NeRF object to get the final camera worldspace matrix using the following equation: ${C_t} = {N}^{-1}{C_o}$ where $C_t$ is the worldspace transformation of the transformed Blender virtual camera, $N$ is the worldspace transformation of the NeRF representation, and $C_o$ is the worldspace transformation of the Blender virtual camera. This allows us to re-position, rotate, and scale the NeRF representation in Blender and animate its transform over the render sequence.

We calculate the FOV of the camera at each frame based on the sensor fit (horizontal or vertical), angle of view, and aspect ratio allowing for dynamic FOV changes. Camera properties in the JSON file are based on user specified fields for the Blender virtual camera such as resolution and camera type (perspective or equirectangular). Finally, we construct the full JSON object and write it to the file path specified by the user.

\begin{figure}[h]
  \centering
  \includegraphics[width=\linewidth]{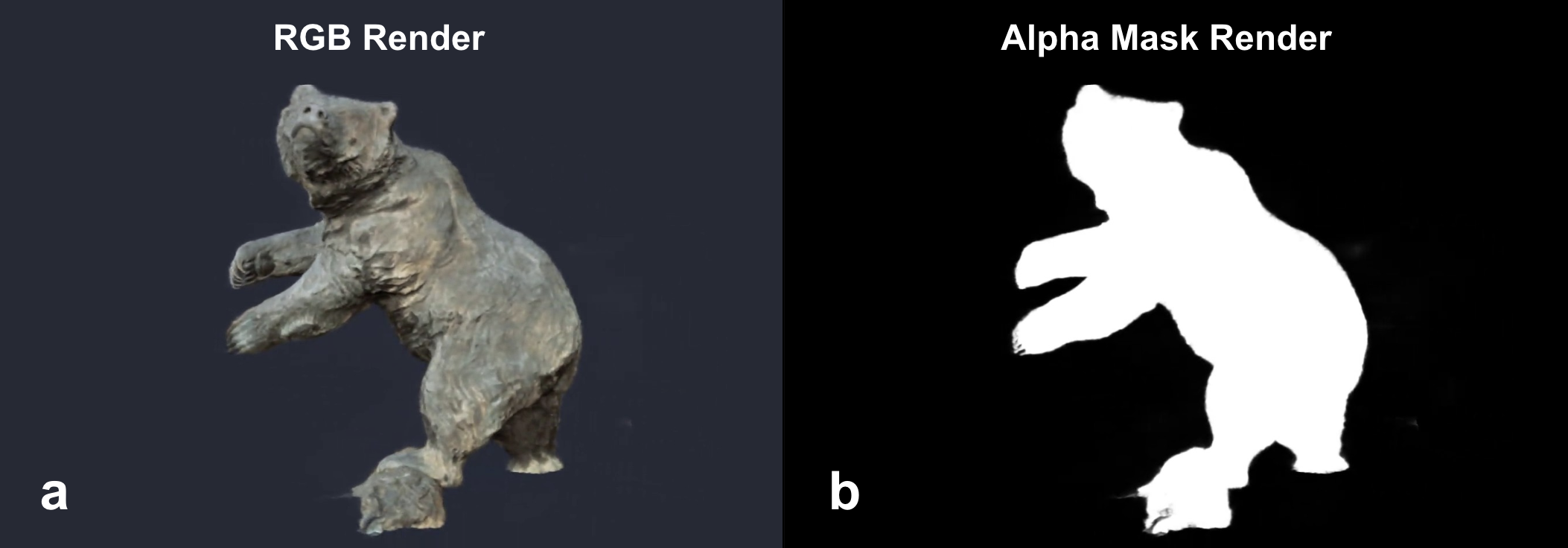}
  \caption{Cropped NeRF object (a) RGB render and (b) accumulation render as an alpha mask}
  \Description{Renders from Nerfstudio to construct cropped NeRF object}
  \label{fig:rgb-acc-render}
\end{figure}

This add-on allows for compositing multiple NeRF objects and environments in a single scene. This is achieved by rendering an RGB render and an accumulation render for each of the cropped NeRF objects individually as shown in Figure~\ref{fig:rgb-acc-render}. The accumulation render from Nerfstudio acts as an alpha mask which can be used to remove the background from the RGB render of the NeRF object in a video editing software. It is possible to simulate shadows cast by NeRF objects over a mesh or NeRF background by compositing in a render of the shadow of the NeRF object mesh representation over a shadow catcher.

\begin{figure}[h]
  \centering
  \includegraphics[width=\linewidth]{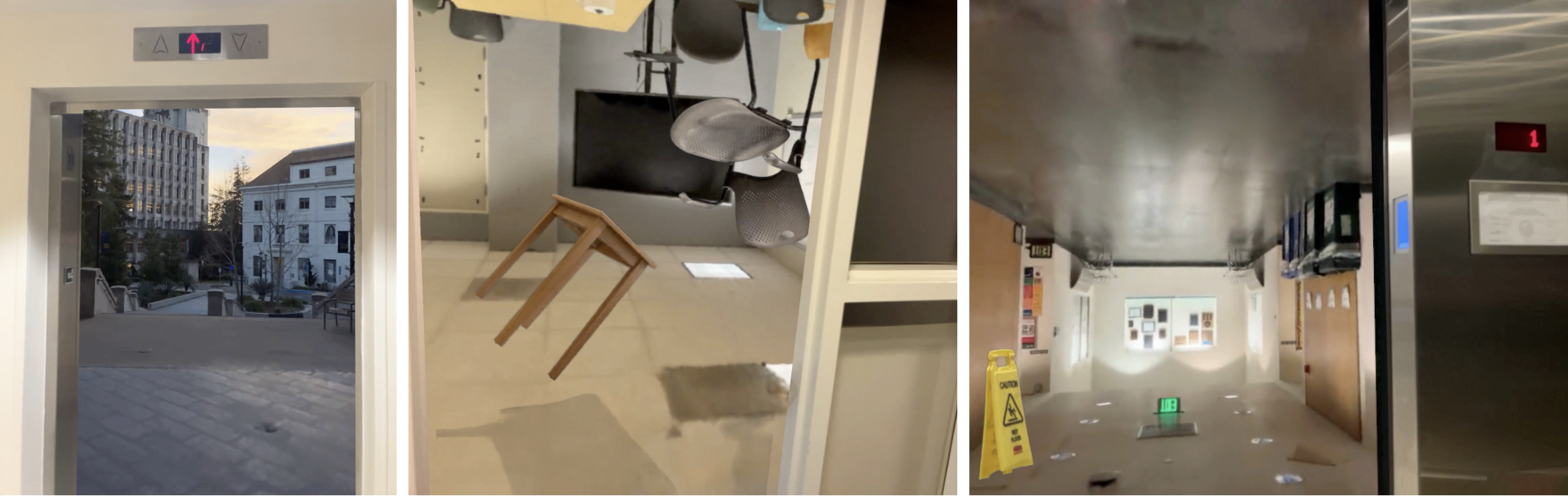}
  \caption{NeRFs composited in live action footage}
  \Description{Composites of NeRF environments and objects motion tracked in real camera footage using the add-on.}
  \label{fig:motion-track}
\end{figure}

We can make use of additional features in Blender such as green screen compositing, motion tracking, and compositing over live-action footage with NeRFs. We can also render an equirectangular 360 image from the NeRF and apply it as an environment map to the Blender scene to relight Blender objects. Our add-on also can rescale the NeRF scene to a real world scale, enabling rendering VR180 or omni-directional stereo VR videos. We showcase the add-on's capabilities by presenting composites featuring portal effects, NeRF objects floating in NeRF environments, and NeRFs composited into real-life footage such as an elevator interior as seen in Figure~\ref{fig:motion-track}.

\section{Conclusion}

Our add-on provides a solution for integrating NeRFs into traditional compositing VFX pipelines using Blender and Nerfstudio. This approach of generating NeRF aligned camera paths can be adapted to other 3D tool sets and workflows, enabling a more seamless integration of NeRFs into visual effects and film production. Future work can explore the use of rendering depth maps of the NeRFs which can improve the quality of compositing of NeRF objects with each other and with other environmental effects. Our contribution offers an exciting avenue for exploring the full potential of NeRFs in visual effects and provides a step towards achieving more realistic and immersive environments for film and other applications.

\begin{acks}
I would like to acknowledge the members of the Nerfstudio team, in particular Matthew Tancik and Angjoo Kanazawa, as well as the Nerfstudio open-source contributors.
\end{acks}

\bibliographystyle{ACM-Reference-Format}
\bibliography{sample-base}

\end{document}